\documentclass[a4paper,twoside]{article}

\usepackage{epsfig}
\usepackage{subcaption}
\usepackage{calc}
\usepackage{amssymb}
\usepackage{amstext}
\usepackage{amsmath}
\usepackage{amsthm}
\usepackage{multicol}
\usepackage{pslatex}
\usepackage{apalike}
\usepackage[bottom]{footmisc}
\usepackage{float}
\usepackage{placeins}
\usepackage{caption}

\usepackage[pagebackref,breaklinks,colorlinks]{hyperref}
\usepackage[capitalize]{cleveref}
\usepackage{booktabs}
\usepackage{graphics}
\crefname{section}{Sec.}{Secs.}
\Crefname{table}{Table}{Tables}
\crefname{table}{Tab.}{Tabs.}
\crefname{equation}{Eq.}{Eqs.}

\usepackage{color, colortbl}
\definecolor{Gray}{gray}{0.95}
\definecolor{Green}{rgb}{0.1,0.9,0.75}

\newcommand{\white}[1]{\textcolor[rgb]{1,1,1}{#1}}

\newcommand{\tb}[1]{\textcolor[rgb]{0,0,0}{#1}} 

\usepackage{SCITEPRESS}     

\begin{document}

\title{\tb{Triple-stream Deep Metric Learning of Great Ape Behavioural Actions}\vspace{-10pt}}

\author{\authorname{Otto Brookes\sup{1}, Majid Mirmehdi\sup{1}, Hjalmar Kühl\sup{2}, Tilo Burghardt\sup{1}}
\affiliation{\sup{1}Department of Computer Science, University of Bristol, United Kingdom}
\affiliation{\sup{2}Evolutionary and Anthropocene Ecology, iDiv, Leipzig, Germany\vspace{5pt}}
\email{otto.brookes@bristol.ac.uk, majid@cs.bris.ac.uk, tilo@cs.bris.ac.uk, hjalmar.kuehl@idiv.de}\vspace{-20pt}
}

\keywords{\tb{Animal Biometrics, Multi-stream Deep Metric Learning, Animal Behaviour, Great Apes, PanAf-500 Dataset}\vspace{-10pt}}


\abstract{\tb{We propose the first metric learning system for the recognition of great ape behavioural actions. Our proposed triple stream embedding architecture works on camera trap videos taken directly in the wild and demonstrates that the utilisation of an explicit DensePose-C chimpanzee body part segmentation stream effectively complements traditional RGB appearance and optical flow streams. We evaluate system variants with different feature fusion techniques and long-tail recognition approaches. Results and ablations show performance improvements of $\sim$~12\% in top-1 accuracy over previous results achieved on the PanAf-500 dataset containing 180,000 manually annotated frames across nine behavioural actions. Furthermore, we provide a qualitative analysis of our findings and augment the metric learning system with long-tail recognition techniques showing that average per class accuracy -- critical in the domain -- can be improved by $\sim$~23\% compared to the literature on that dataset. Finally, since our embedding spaces are constructed as metric, we provide first data-driven visualisations of the great ape behavioural action spaces revealing emerging geometry and topology. We hope that the work sparks further interest in this vital application area of computer vision for the benefit of endangered great apes. We provide all key source code and network weights alongside this publication.\vspace{-20pt}}}

\onecolumn \maketitle \normalsize \setcounter{footnote}{0}\vfill

\begin{minipage}[c]{\textwidth}
\centering
\includegraphics[width=390pt,height=190pt]{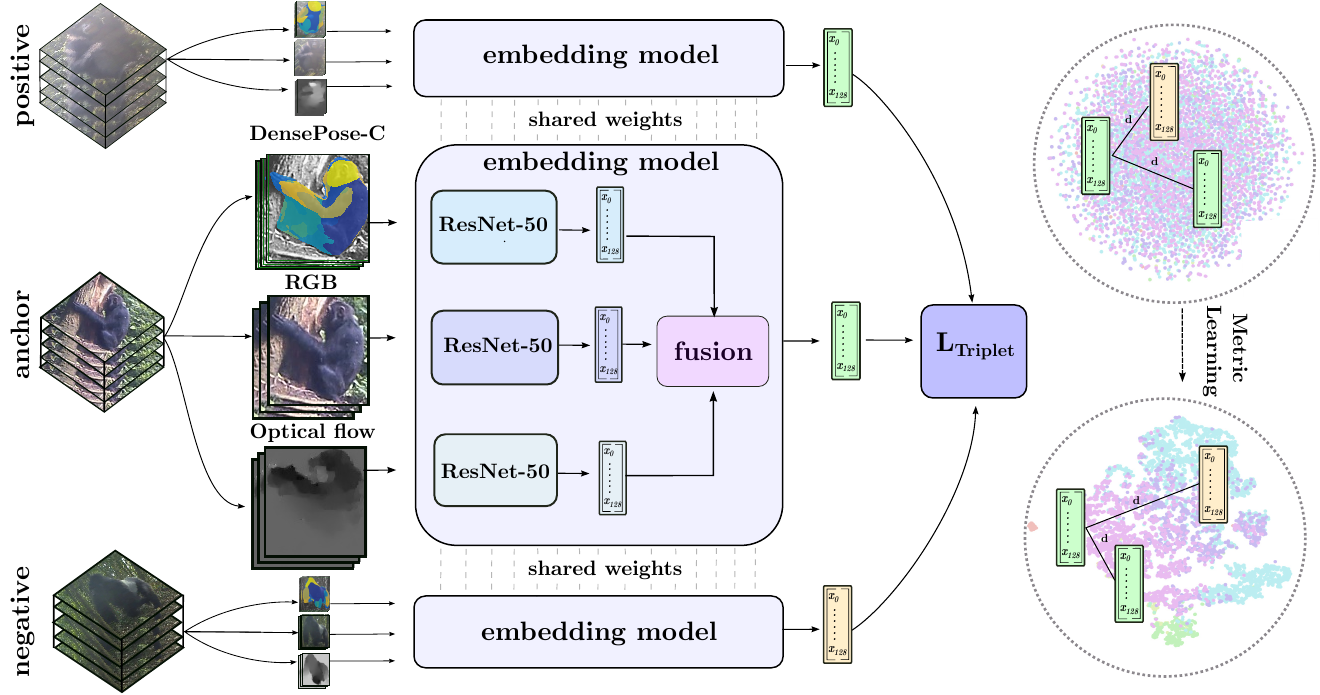}
\captionof{figure}{\tb{\textbf{System Overview}. {\small Our proposed triple-stream metric learning approach utilises all RGB appearance, optical flow, and DensePose-C segmentations of chimps in videos. Exploiting hybrid reciprocal triplet and cross entropy losses, the model is then trained to map embeddings representing great ape behavioural actions onto a metric space, where semantically similar representations are geometrically close forming natural clusters. This pipeline improves on state-of-the-art classification performance and allows for visualisations of the underpinning space of behavioural actions.  (best viewed zoomed)}}}
\label{fig:overview_diagram}
\end{minipage}


\section{\uppercase{Introduction}}
\label{sec:intro}

\tb{As the climate crisis gathers pace, the threat to many endangered species grows ever more perilous~\cite{almond2022living}. All species of great apes are, for instance, listed as endangered or critically endangered according to the IUCN Red List~\cite{IUCN}}
\newpage
\tb{\white{..}\vspace{250pt} \white{. . . . . . . . . . there is urgent need for methods} Consequently, there is urgent need for methods that can help to monitor population status and assess the effectiveness of conservation interventions~\cite{kuhl2013animal,congdon2022future,tuia2022}. This includes the recognition of behaviors and variation therein, as an integral part of biological diversity~\cite{dominoni2020conservation,carvalho2022using}.}

\tb{Previous works have employed deep neural networks which leverage multiple modalities, such as RGB, optical flow, and audio~\cite{sakib2020visual,bain2021automated}, for the classification of great ape behaviours and actions. However, higher level abstractions such as \textit{pose} or \textit{body part} information have remained unexplored for addressing this task. In response, we propose utilising the latter \textit{together} with RGB and optical flow in a triple-stream metric learning system (see Fig.~\ref{fig:overview_diagram}) for improved classification results and domain visualisations relevant to biologists.}

\begin{figure}[ht]
\centering
\includegraphics[width=\linewidth,height=300pt]{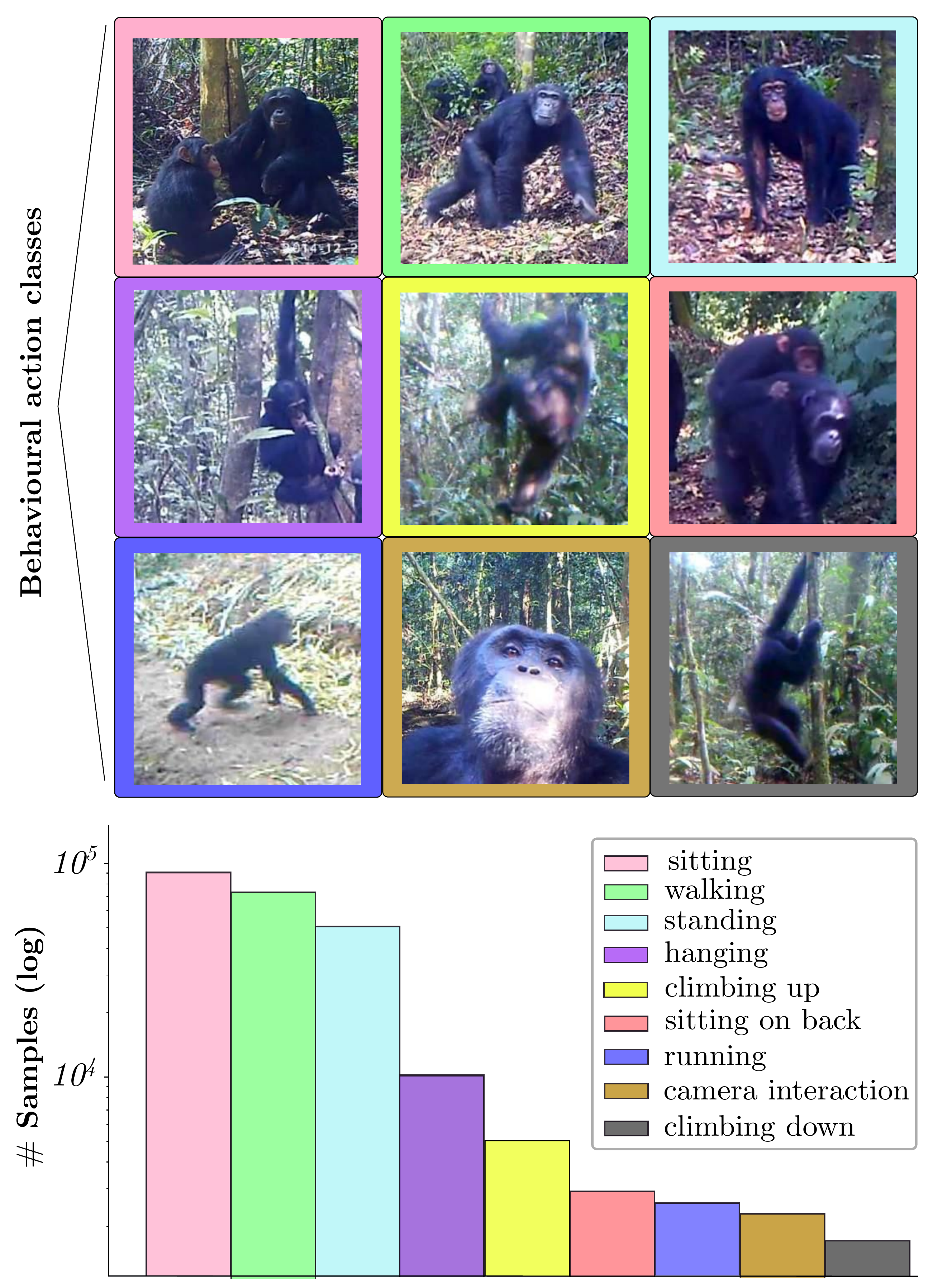}
\caption{\tb{\textbf{Behavioural Actions in the PanAf-500 Data}. \textmd{Examples of each one of the nine behavioural action classes (\textit{top}) and their distribution across the approx. 180k frames in the dataset (\textit{bottom}). Note the imbalance of two orders of magnitude in the distribution.  (best viewed zoomed)}}}
\label{fig:human_class_dist}
\end{figure}

\tb{{\bf Great Ape Activities - } This paper will focus on \emph{great ape activity recognition}, where the coarse activity classes used are illustrated in Fig.~\ref{fig:human_class_dist} for the utilised PanAf-500 dataset (see Sec.~\ref{SecData}). Note that computer vision would traditionally categorise these classes as actions whilst in the biological realm they represent behaviour (or aspects thereof) often captured in ethograms~\cite{nishida1999ethogram,zamma_matsusaka_2015}. For clarity, in this paper we will refer to these classes as \emph{behavioural actions} recognising historical traditions in both disciplines.}

\tb{We will approach the classification task via a deep \textit{metric} learning system~\cite{karaderi2022visual} that embeds inputs into a latent space and uses geometric distances to form distributions that align with the semantic similarity captured by the classes~\cite{hermans2017defense,musgrave2020pytorch}. A major advantage over standard supervised systems is that sample distances in visualisations of the latent space always relate to learned similarity and, thus, are more naturally interpretable by experts. We will also analyse the role that additional DensePose-Chimp information~\cite{sanakoyeu2020transferring} can play in improving recognition performance compared to systems that utilise RGB and optical flow only. Lastly, as shown by Sakib and Burghardt~\cite{sakib2020visual}, there are significant challenges in correctly classifying behavioural actions which occur infrequently and form the distribution tail (see Fig.~\ref{fig:human_class_dist}). To address this, we will employ three long-tailed recognition~(LTR) techniques to improve performance on tail classes;~(i) logit adjustment~\cite{menon2020long};~(ii) class balanced focal loss~\cite{cui2019class}; and (iii)~weight balancing~\cite{alshammari2022long}}. 

\tb{In summary, our contributions are as follows: (i)~we implement the first deep \emph{metric} learning system for recognising great ape behavioural actions; (ii)~we show that utilising explicit pose information has a significant positive effect on recognition performance in this domain; and (iii)~we establish that existing LTR techniques can be applied in a metric learning setting to improve performance on tail classes for the problem. The proposed approaches improve the state-of-the-art performance benchmarks with respect to top-1 ($\sim$~85\%) and average per class ($\sim$~65\%) accuracy on the PanAf-500 dataset.\vspace{-15pt}}


\section{\uppercase{Related Work}\vspace{-5pt}}
\tb{Action recognition aims to classify actions observed in video~\cite{kalfaoglu2020late,shaikh2021rgb}. Learning spatio-temporal features characteristic for actions~\cite{simonyan2014two} via various deep learning paradigms forms the approach of choice in the domain of human action recognition (HAR). We will briefly review concepts from this field, before discussing specifc relevant great ape behavioural action recognition and LTR methods.}

\tb{{\bf Human Action Recognition - }
Although there are numerous deep learning approaches to action recognition~\cite{zhou2018temporal,lin2019tsm,tran2019video,kalfaoglu2020late,pan2019compressing,majd2020correlational,sharir2021image,zhang2021vidtr} this work focuses on multi-stream architectures, which address key aspects of the action recognition problem (e.g., spatial and temporal) independently and explicitly. Feichtenhofer et al.~\cite{feichtenhofer2019slowfast} introduced the SlowFast architecture which employs two streams, each operating at different frame rates; a slow, low frame-rate pathway captures spatial information while the fast, high frame-rate pathway captures fine temporal detail. Other types of multi-stream networks process different visual modalities. Simonyan~\cite{simonyan2014two} introduced a two-stream network that processes RGB and optical flow to exploit spatial and temporal semantics, respectively. Since then, several networks that utilise additional modalities, such as motion saliency~\cite{zong2021motion} and audio~\cite{wang2021multi}, have been introduced. Recently, the introduction of pose, which is critical for the perception of actions~\cite{le2022comprehensive}, has shown promising results in multi-stream architectures~\cite{hong2019contextual,hayakawa2020recognition,duan2021ntu,li2022perf}. In particular, the DensePose format provides an opportunity to exploit fine-grained, segmentation map-based pose representations for action recognition. Hayakawa et al.~\cite{hayakawa2020recognition} combine RGB and DensePose estimations in a two-stream network and demonstrate strong performance on egocentric footage of humans. Whilst such significant progress has been made in the domain of HAR, research into great ape behavioural action recognition is still in its infancy and few systems have been tested on natural datasets.}

\tb{{\bf Great Ape Domain - } To date, two systems have attempted automated great ape behavioural action recognition, both are multi-stream architectures. The first~\cite{sakib2020visual} is based on the two-stream convolutional architecture by Simonyan et al.~\cite{simonyan2014two} and used 3D ResNet-18s for feature extraction and LSTM-based fusion of RGB and optical flow features. They report top-1 accuracy of 73.52\% across the nine behavioural actions in the PanAf-500 dataset (see Sec.~\ref{SecData}) and a relatively low average per class accuracy (42.33\%), highlighting the issue of tail class performance. The second, proposed by Bain et al.~\cite{bain2021automated}, is a deep learning system that requires both audio and video inputs and detects two specific behaviours; buttress drumming and nut cracking. Their system utilised a 3D ResNet-18 and a 2D ResNet-18 for extraction of visual and assisting audio features, respectively, in different streams. They achieved an average precision of 87\% for buttress drumming and 85\% for nut cracking on their unpublished dataset. However, the multi-modal method is not applicable to all camera trap settings since many older models do not provide audio. It cannot be utilised on the PanAf-500 dataset since many clips there do not contain audio.}

\tb{{\bf Long-tailed Recognition - } Most natural recorded data exhibits long-tailed class distributions~\cite{liu2019large}. This is true of great ape camera-trap footage which is dominated by commonly occurring behaviours - even with only the nine classes of the PanAf-500 data the distribution shows a clear tail~(see Fig.~\ref{fig:human_class_dist}). Without addressing this issue, models trained on such data often exhibit poor performance on rare classes. Various counter-measures have been proposed~\cite{verma2018manifold,kang2019decoupling,zhang2021bag}. Class balanced losses assign additional weights, typically determined by inverse class frequencies, to samples from rare classes and have yielded strong results when coupled with techniques to reduce per-class redundancy~\cite{cui2019class}. Similarly, logit adjustment uses class frequencies to directly offset output logits in favour of minority classes during training~\cite{menon2020long}. An orthogonal approach, based on the observation that weight norms for rare classes are smaller in naively trained classifiers, is to perform weight balancing~\cite{alshammari2022long}. These techniques have achieved strong results on several LTR benchmarks.} 

\tb{Before detailing how we use triple-stream metric learning with explicit DensePose-Chimp processing and LTR extensions for behavioural action recognition, we will briefly outline the utilised dataset.\vspace{-15pt}}


\section{\uppercase{Dataset}\vspace{-5pt}}
\label{SecData}
\tb{The \textit{Pan-African} dataset, gathered by the Pan African Programme: ‘The Cultured Chimpanzee’, comprises $\sim$~20,000 videos from footage gathered at 39 study sites spanning 15 African countries. Here we utilise a 500 video subset, PanAf-500, specifically ground-truth labelled for use in computer vision under reproducible and comparable benchmarks. It includes frame-by-frame annotations for full-body locations of great apes and nine behavioural actions~\cite{sakib2020visual} across approximately 180k frames (see.~\cref{fig:manual_annotations}). ~\cref{fig:human_class_dist} displays the behavioural actions classes in focus together with their distribution. We utilised the PanAf-500 dataset for all experiments and employ the same training and test partitions described in~\cite{sakib2020visual}.\vspace{-15pt}}

\section{\uppercase{Method}\vspace{-5pt}}
\label{sec:method}
\tb{The proposed system utilises three visual modalities as input; RGB, optical flow, and DensePose-C estimations~\cite{sanakoyeu2020transferring}, as illustrated in~Fig.~\ref{fig:overview_diagram}). All optical flow images are pre-computed using OpenCV's implementation of the Dual TV L1 algorithm~\cite{zach2007duality}. We employ the model developed by Sanakoyeu et al.~\cite{sanakoyeu2020transferring} to generate DensePose-C segmentations describing chimpanzee pose. The model predicts dense correspondences between image pixels and a 3-D object mesh where each mesh represents a chimpanzee body part specified by a selector $I$ and local surface coordinates within each mesh indexed by $U$ and $V$. Frame-by-frame application to each of the PanAf-500 videos yields DensePose-C estimates expressed in $IUV$ coordinates.\vspace{10pt}}

\setcounter{figure}{2}
\begin{figure}[t]
\begin{minipage}{\columnwidth}
    \includegraphics[width=0.49\linewidth]{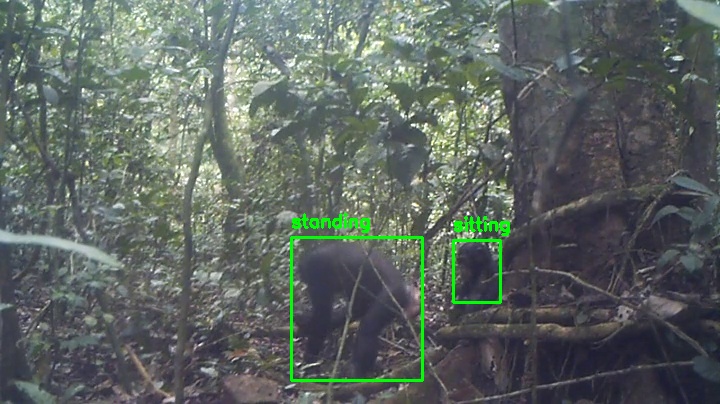}
    \includegraphics[width=0.49\linewidth]{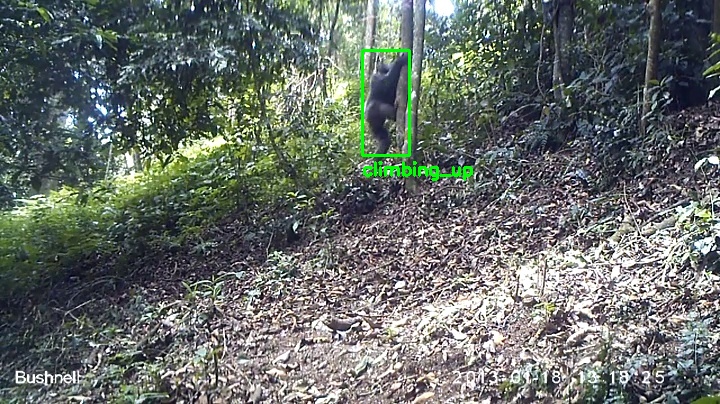}
\begin{minipage}{\columnwidth}
    \includegraphics[width=0.49\linewidth]{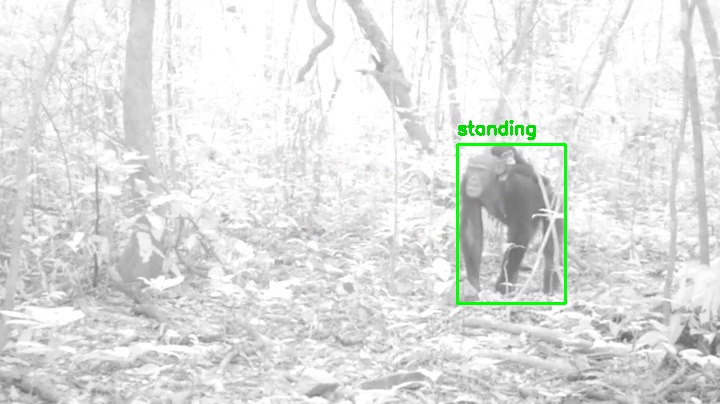}
    \includegraphics[width=0.49\linewidth]{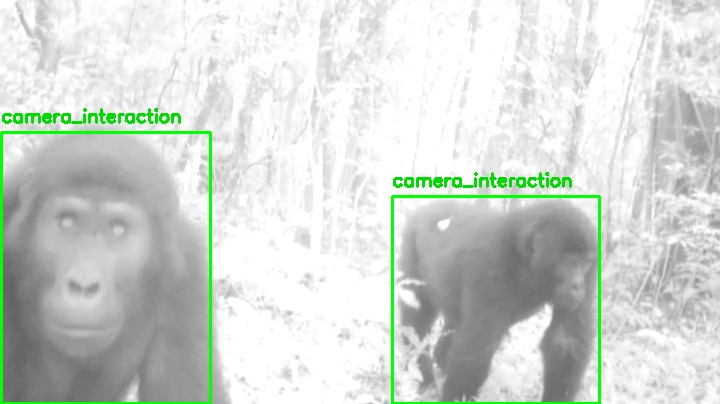}
    \end{minipage}
\end{minipage}
\caption{\tb{\textbf{Frame-by-frame Ground Truth Annotations}. \textmd{Four still frames from PanAf-500 videos with annotations of location  (green boxes) and behavioural actions (visualised as text) of the apes in-frame. (best viewed zoomed)}}}\label{fig:manual_annotations}
\end{figure}\vspace{-10pt}


 \setcounter{figure}{3}
\begin{figure}[t]
    \centering
    \includegraphics[width=210pt,height=250pt]{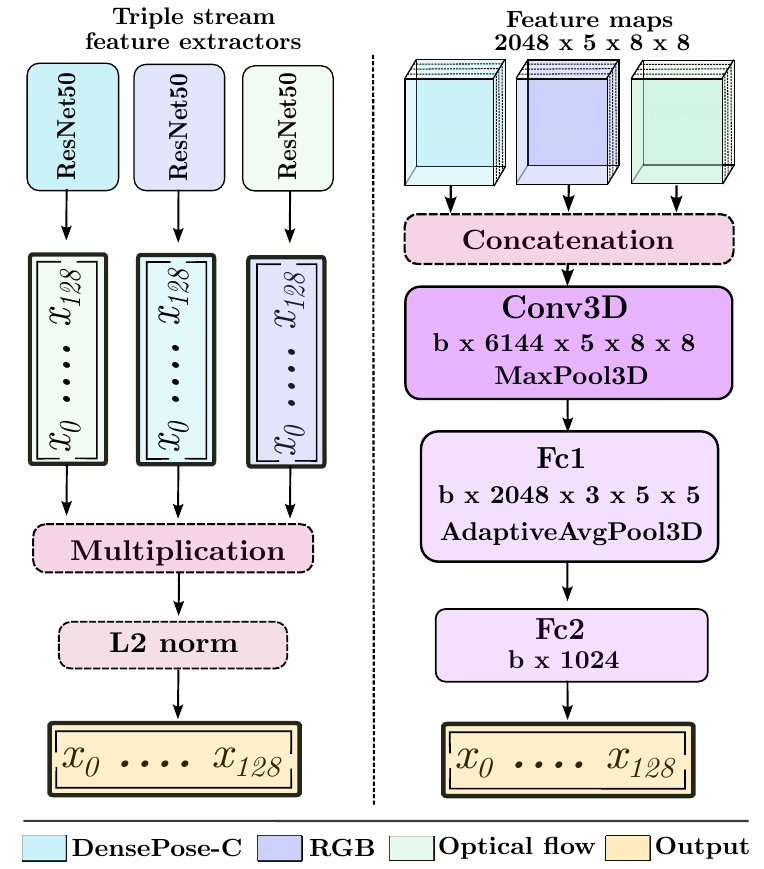}
    \caption{\tb{\textbf{Fusion Head Schematics}. \textmd{A component break-down of fusion by element-wise multiplication (\textit{left}) and convolutional fusion (\textit{right}) as applied for our work to explore their impact on performance.}}}
    \label{fig:fusion_head}
\end{figure}

\tb{Each of the three input modalities is fed into a 3D ResNet-50~\cite{du2017closer} backbone, which together act as a feature extractor (see Fig.~\ref{fig:overview_diagram}). The input tensors into the backbones are 3D since inputs are processed in snippets, that is each stream accepts a sequence of $n$ consecutive RGB frames, optical flow images, or {$IUV$} coordinates, respectively. The final fully-connected layer outputs an $n$-dimensional encoding for each stream. These are fused into a single embedding using three popular approaches; (i) simple averaging across streams; (ii) convolutional fusion whereby stream features are concatenated and passed to a 3D convolutional layer as a volume; and (iii) element-wise multiplication of all three embedding vectors followed by $L2$ normalisation. The latter two approaches are illustrated in~detail in~\cref{fig:fusion_head}. A linear layer at the end of the fusion head finally outputs the unified embedding as logits. Whilst this system was trained via metric learning - visually sketched in Fig.~\ref{fig:overview_diagram} (right) - a $k$-NN classifier is used to perform inference in the embedding space during evaluation.}

\tb{Let the parameters of this network~$f_{\theta}(\cdot)$ be denoted by~$\theta$. Furthermore, let $f_{\theta}(x)~{=}~x$~be the shorthand for referring to embeddings. Our metric learning objective is, thus, to minimise the distance between anchor-positive embedding pairs~$d(x_{a}, x_{p})$~and
maximise distance between anchor-negative embedding pairs~$d(x_{a}, x_{n})$, where $d$ represents a Euclidean. Instead of using standard triplet loss~\cite{hermans2017defense}~$L_{TL}$, we use an improved version~\cite{andrew2021visual}, where the model is optimised via a hybrid reciprocal triplet and softmax cross-entropy loss:}
\tb{\begin{equation}
\label{eq:hybrid}
L_{RC} = L_{CE} + \lambda~L_{RT}.
\end{equation}
It is assembled from two components balanced by $\lambda=0.1$ as given in~\cite{andrew2021visual}. The two components themselves are evaluated as:
\begin{equation}
\label{eq:rtl}
L_{RT} = d(x_{a}, x_{p})+\frac{1}{d(x_{a}, x_{n})}
\end{equation}
\begin{equation}
\label{eq:ce}
L_{CE} = -\log\left(\frac{e^{x_{y}}}{\sum^{C}_{i=1}e^{x_{i}}}\right),
\end{equation}
where $C$ denotes the total number of classes and $y$ are the class labels.}

\tb{In order to extend this system into the LTR domain we substitute the softmax cross-entropy term for losses calculated using; (i)~cross-entropy softmax with logit adjustment~\cite{menon2020long}~$L_{LA}$; (ii)~class-balanced focal loss~\cite{cui2019class}~$L_{CB}$; and (iii)~class-balanced focal loss with weight balancing~\cite{alshammari2022long}.
The first two losses are evaluated as follows:
\begin{equation}
\label{eq:logit_adjusted}
L_{LA} = -\log\left(\frac{e^{x_{y}}+\tau~{\cdot}~\log~\pi_{y}}{\sum^{C}_{i=1}e^{x_{i}+\tau~{\cdot}~\log~\pi_{i}}}\right),
\end{equation}
\begin{equation}
\label{eq:class_balanced}
L_{CB} = -~\frac{1-\beta}{1-\beta^{n_{y}}}\sum^{C}_{i=1}(1-p_{i})^{\gamma}~\log(p_{i}),
\end{equation}
where $\pi$ represents the class priors (i.e., class frequencies in the training set) and temperature factor $\tau=1$, $\beta=0.99$ is the re-weighting hyper-parameter, $n$ is the total number of samples, $y$ are the classes, $\gamma=1$ is the focal loss hyper-parameter and $p_{i} = {\sigma(x_{i})}$.
Balancing the network weights $\theta$ is performed via a MaxNorm constraint~${\parallel}{\theta_{l,i}}{\parallel}^{2}_{2} \leq \delta^{2}, \forall{i}$ given in~\cite{alshammari2022long} imposed on each class filter $i$ in the last layer $l$ of the network where $\delta$ is the L2-norm ball radius. We will reference a $L_{CB}$-based optimisation where weight balancing is performed via~$L_{WB}$.}

\tb{Methodologically, this described architecture approaches the learning of behavioural great ape actions via five key capabilities: 1) utilisation of multiple relevant input modalities across an entire video snippet; 2) effective streamed content encoding; 3) fusion into a single embedding space; 4) metric space optimisation so that distances naturally reflect semantic similarity; and 5) taking into account class imbalances common to the domain content.\vspace{-15pt}}

\section{\uppercase{Experiments}\vspace{-5pt}}

\subsection{General Training Setup\vspace{-5pt}}

\tb{We train our architecture via SGD optimisation using batch size 32 and learning rate $10^{-4}$. Feature extractor backbones are initialised with Kinetics-400~\cite{kay2017kinetics} pre-trained weights and training runs are distributed over 8 Tesla V100 GPUs for 100 epochs.\vspace{-5pt}}

\subsection{Baselines and Stream Ablations\vspace{-5pt}}
\tb{As shown in Tab.~\ref{tab:main_results}, we first establish performance benchmarks for one and two stream baseline architectures of our system~(rows 2--5) against the current state-of-the-art~(row 1), which uses a ResNet-18 backbone with focal loss~$L_{FL}$, SGD, and LSTM-based frame fusion~\cite{sakib2020visual}. As expected, we confirmed that - using identical setups and losses - adding an optical flow stream is beneficial in the great ape domain mirroring HAR results~(see rows 2 vs 4, and 3 vs 5). Additionally, models trained using $L_{RC}$ consistently outperformed standard triplet loss $L_{RC}$ scenarios~(see rows 2 vs 3, and 4 vs 5). Finally, a dual-stream version of our proposed architecture trained with $L_{RC}$ outperforms the state-of-the-art by a small margin~(see rows 1 vs 5).\vspace{-5pt}}

\subsection{Triple-Stream Recognition\vspace{-5pt}}

\tb{As given in Tab.~\ref{tab:main_results} rows 6--8, our proposed triple-stream architecture significantly outperforms all baselines with regards to top-1 accuracy, achieving up to $85.86\%$. Thus, explicit DensePose-C information appears a useful information source for boosting behavioural action recognition in great apes. However, without LTR techniques all our triple-stream models are significantly outperformed by a dual-stream setting (row 5) with regards to average per-class accuracy. This reduction is caused by significantly poorer performance on minority classes~(see Sec.~\ref{SecFusion}). }

\begin{table}[t]
\footnotesize
  \centering
    \caption{\textmd{\tb{\textbf{Behavioural Action Recognition Benchmarks.} Top-1 and average per-class (C-Avg) accuracy performance on the PanAf-500 dataset for the current state-of-the-art~(row 1), single and dual-stream baselines~(rows 2--5), and our triple-stream networks~(rows 6--8) for different fusion methodologies and losses tested.}}}
  \resizebox{\columnwidth}{!}{%
  \begin{tabular}{r*{5}{l}}
    \toprule
    \multicolumn{2}{l}{\textbf{Models/Streams}} & \textbf{Fusion} & \textbf{Loss} & \textbf{Top-1} & \textbf{C-Avg} \\
    \midrule
    \hline
    & & & &  \\
     \multicolumn{2}{l}{\textbf{Sakib et al. 2020}} & & & & \\
     1  & $RGB{+}OF$ & LSTM & $L_{FL}$ & \textbf{73.52\%} & \textbf{42.33\%}  \\
     \hline
     & & & &  \\
    \multicolumn{2}{l}{\textbf{Up to Dual-Stream}} & & & & \\
     2 & $RGB$ $only$ & None & $L_{TL}$ & 55.50\% & 32.67\% \\
     3 & $RGB$ $only$ & None & $L_{RC}$ & 74.24\% & 55.76\% \\
     4 & $RGB{+}OF$ & Avg & $L_{TL}$ & 62.90\% & 39.10\% \\
     5 & $RGB{+}OF$ & Avg & $L_{RC}$ & \textbf{75.02\%} & \underline{\textbf{61.97\%}} \\
    \hline
    & & & &  \\
    \multicolumn{2}{l}{\textbf{Triple-Stream {(Ours)}}} & & & & \\
     6 & $RGB{+}OF{+}DP$ & Avg & $L_{RC}$ & 81.71\% & 46.61\% \\
    7 & $RGB{+}OF{+}DP$ & Conv & $L_{RC}$ & 82.04\% & \textbf{56.31\%} \\
    8 & $RGB{+}OF{+}DP$ & Elem & $L_{RC}$ & \underline{\textbf{85.86\%}}  & 50.50\% \\
    \bottomrule
  \end{tabular}
  }
  \label{tab:main_results}
\end{table}

\setcounter{figure}{4}
\begin{figure*}[ht]
    \centering
    \includegraphics[width=\textwidth, scale=0.8]{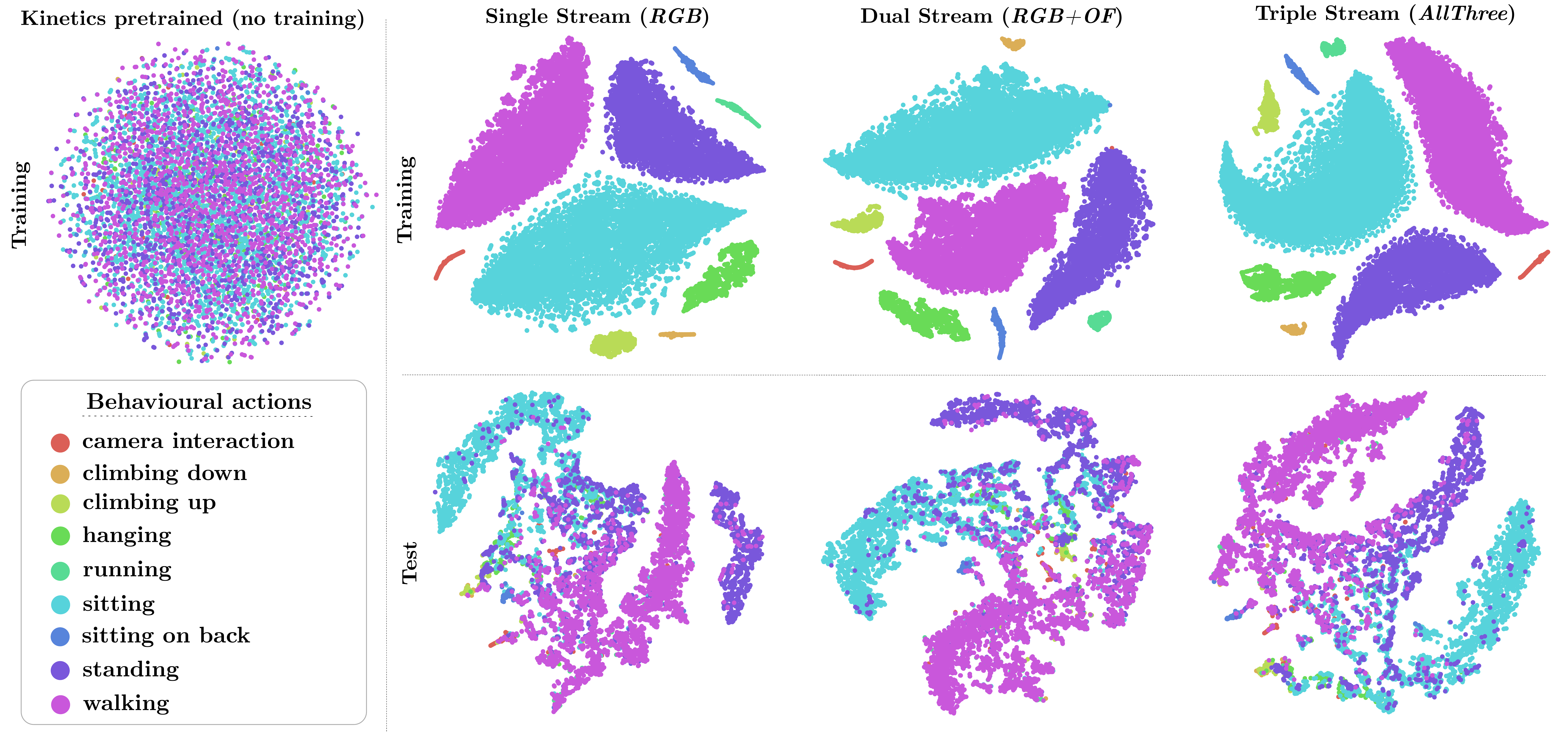}
    \caption{\textbf{Visualisations of Great Ape Behavioural Action Spaces}. \textmd{A 2D t-SNE~\cite{wattenberg2016how} visualisation of the 128-dimensional training (top-right) and test (bottom-right) embeddings produced by the single, dual and three-stream network with convolutional fusion. We can see that training set embeddings from all classes are clustered cleanly. In contrast, test set embeddings show significant overlap and only embeddings from majority classes form distinct clusters. This is consistent with the high top-1 accuracy and relatively low average per-class accuracy reported in Tab.~\ref{tab:main_results}}}
    \label{fig:all_embeddings}
\end{figure*}

\tb{Since the learned behavioural action embeddings are constructed as metric from the outset, they can be visualised meaningfully -- we note that such data-driven visualisations are novel in the primatology domain. \cref{fig:all_embeddings} depicts such learned spaces for our data and architecture where, independent of stream cardinality, embeddings cluster the training data cleanly. This is of course expected given above $99\%$ top-1 \emph{training} accuracy in all settings. Yet, behavioural actions of great apes are highly intricate as well as variable and, even with approx. $144,000$ training frames used, the model clearly shows signs of overfitting. As a result, test set embeddings exhibit significant cluster overlap. Sample groups representing sitting, standing, and walking, for instance, blend into one another. In addition to overfitting, this also highlights the transitional nature of these often temporarily adjacent and smoothly changing actions. Thus, future temporally transitional ground truth labelling may be needed to represent behavioural great ape action in the PanAf-500 dataset more authentically.\vspace{-5pt}}

\subsection{Fusing Streams\vspace{-5pt}}
\label{SecFusion}
\tb{When looking at the impact of information fusion methods on performance in more detail, we find that benchmarks vary significantly (see~Tab.~\ref{tab:main_results} rows 6--8) when we test averaging, element-wise multiplication, and convolutional fusion, as described in \cref{sec:method}. Results show that convolution and element-wise multiplication improve performance slightly across both metrics when compared with averaging: top-1 accuracy improves by $0.33\%$ and $4.1\%$, respectively (see rows 6--8). However, the most significant gains are observed with respect to average per class accuracy which increases by 3.44\% for element-wise multiplication and 9.7\% for convolutional fusion. Learnable parameters in the convolution method clearly help blending information even when only fewer samples are available for training. Building on this improvement, we will next investigate the impact of LTR methods in order to benefit tail class performance.\vspace{-5pt}}

\subsection{Long-tail Recognition\vspace{-5pt}}

\begin{figure*}[!ht]
    \centering
    \includegraphics[width=\textwidth]{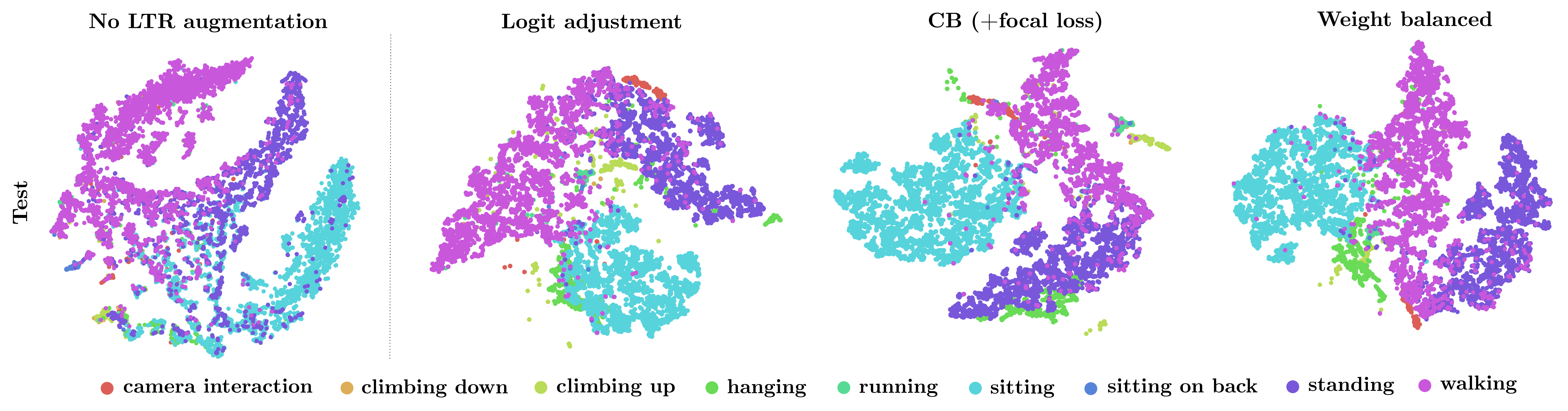}
    \caption{\textbf{Long-tail Test Embeddings}. \textmd{A 2D t-SNE visualisation of the 128-dimensional test embeddings produced by the three-stream network with convolutional fusion alone (leftmost) and augmented with each LTR technique;~(i)~logit adjustment (ii)~CB (+focal loss) and (iii)~weight balancing. All LTR-augmented methods improve clustering of embeddings belonging to tail classes. They appear more clearly separated and exhibit less overlap when compared with the non-LTR method.}}
    \label{fig:long_tail_embeddings}
\end{figure*}

\tb{When grouping behavioural actions into \textit{head} (covering sitting, standing, and walking) and remaining \textit{tail} classes based on frequency in the data~(see Fig.~\ref{fig:human_class_dist}), a significant performance gap becomes apparent even when using the so far best C-Avg performing model~(see Tab.~\ref{tab:long_tail} row 1). Employing LTR techniques can, however, reduce this gap and improve average per-class accuracy further as quantified across rows 2--4 in Tab.~\ref{tab:long_tail}). \cref{fig:long_tail_embeddings} shows t-SNE visualisations of the three LTR triple-stream approaches when trained with convolutional feature fusion. Particularly for the class-balanced approaches and weight-balancing setups~(two rightmost), \textit{tail} class clusters appear more clearly separated and class overlap is generally reduced. Thus, for the great ape domain underrepresented classes are indeed an effective source of information for improving action separability in general.\vspace{-15pt}}

\section{\uppercase{Conclusion}\vspace{-5pt}}
\tb{In this work we introduced the first deep metric learning system for great ape behavioural action recognition. We demonstrated that the proposed triple-stream architecture can provide leading state-of-the-art performance when tested on the PanAf-500 camera trap dataset covering 180,000 annotated frames across 500 videos taken in the wild. We demonstrated that the addition of a DensePose-C chimpanzee pose estimation stream into the embedding architecture is highly effective and leads to system performance of 85.86\% top-1 accuracy on the data. We also showed that adding LTR techniques that address poor tail class performance to the system can improve the average per-class accuracy to 65.66\% on the dataset. Despite these improvements we note that both larger annotated datasets to counteract overfitting as well as more temporally blended forms of annotation (e.g. action transition annotations) would benefit the  authenticity of data-driven great ape behavioural representations. We hope that the research presented here sparks further interest in this vital application area for the benefit of endangered species such as great apes.\vspace{-15pt}}
\begin{table}[b]
\footnotesize
  \centering
    \caption{\textmd{\tb{\textbf{LTR-enabled Behavioural Action Recognition Benchmarks.} Average per-class accuracy for our triple-stream network with convolutional fusion for best performing non-LTR method~(row1), and three LTR approaches~(rows 2--4) targetting poor tail class performance.}}}
  \resizebox{\columnwidth}{!}{%
  \begin{tabular}{r*{5}{l}}
    \toprule
    \multicolumn{2}{l}{\textbf{Method/Loss}}  &  \textbf{C-Avg} & \textbf{Head} & \textbf{Tail} \\
    \midrule
    \midrule
        & & & &  \\
    \multicolumn{2}{l}{\textbf{Non-LTR Triple-Stream}} & & &  \\
    1 & $L_{RC}$ &  56.31 & 80.57 & 44.78 \\
    \hline
            & & & &  \\
    \multicolumn{2}{l}{\textbf{LTR Triple-Stream}} & & &  \\
    2 & $L_{LA}$ &   61.76 & \textbf{83.22} & 50.7 \\
    3 & $L_{CB}$ &   63.56 & 77.60 & 55.95 \\
    4 & $L_{WB}$ &   \textbf{65.66} & 82.55 & \textbf{56.26} \\
    \bottomrule
  \end{tabular}
  }
  \label{tab:long_tail}
\end{table}

\section*{\uppercase{\large{Acknowledgements}}\vspace{-5pt}}
\label{SecAck}
{\footnotesize We thank the Pan African Programme: ‘The Cultured Chimpanzee’ team and its collaborators for allowing the use of their data for this paper. We thank Amelie Pettrich, Antonio Buzharevski, Eva Martinez Garcia, Ivana Kirchmair, Sebastian Schütte, Linda Gerlach and Fabina Haas. We also thank management and support staff across all sites; specifically Yasmin Moebius, Geoffrey Muhanguzi, Martha Robbins, Henk Eshuis, Sergio Marrocoli and John Hart. Thanks to the team at https://www.chimpandsee.org particularly Briana Harder, Anja Landsmann, Laura K. Lynn, Zuzana Macháčková, Heidi Pfund, Kristeena Sigler and Jane Widness. The work that allowed for the collection of the dataset was funded by the Max Planck Society, Max Planck Society Innovation Fund, and Heinz L. Krekeler. In this respect we would like to thank: Ministre des Eaux et Forêts, Ministère de l'Enseignement supérieur et de la Recherche scientifique in Côte d’Ivoire; Institut Congolais pour la Conservation de la Nature, Ministère de la Recherche Scientifique in Democratic Republic of Congo; Forestry Development Authority in Liberia; Direction Des Eaux Et Forêts, Chasses Et Conservation Des Sols in Senegal; Makerere University Biological Field Station, Uganda National Council for Science and Technology, Uganda Wildlife Authority, National Forestry Authority in Uganda; National Institute for Forestry Development and Protected Area Management, Ministry of Agriculture and Forests, Ministry of Fisheries and Environment in Equatorial Guinea. This work was supported by the UKRI CDT in Interactive AI under grant EP/S022937/1.}

\bibliographystyle{apalike}
{\small
\bibliography{example}}

\end{document}